%% file: bt_estimation_acc2021.tex
\documentclass[letterpaper, 10 pt, conference]{ieeeconf}  

\IEEEoverridecommandlockouts                              

\usepackage{graphics, cite} 
\usepackage{epsfig} 
\usepackage{amsmath} 
\usepackage{amssymb}  

\usepackage{tikz} 
\usepackage{pgfplots}
\pgfplotsset{compat=1.17}
\usepackage{subcaption}

\usetikzlibrary{arrows}
\usetikzlibrary{arrows.meta}
\usetikzlibrary{calc}
\usetikzlibrary{decorations.markings}
\usetikzlibrary{decorations.pathreplacing}
\usetikzlibrary{decorations.pathmorphing}
\usetikzlibrary{fadings}
\usetikzlibrary{intersections} 
\usetikzlibrary{patterns}

\tikzstyle{spring}=[decorate,decoration={zigzag,pre length=0.3cm,post length=0.3cm,segment length=6}]
\tikzstyle{damper}=[decoration={markings,  
  mark connection node=dmp,
  mark=at position 0.5 with 
  {
    \node (dmp) [inner sep=0pt,transform shape,rotate=-90,minimum width=10pt,minimum height=3pt,draw=none] {};
    \draw ($(dmp.north east)+(2pt,0)$) -- (dmp.south east) -- (dmp.south west) -- ($(dmp.north west)+(2pt,0)$);
    \draw ($(dmp.north)+(0,-4pt)$) -- ($(dmp.north)+(0,4pt)$);
  }
}, decorate]

\title{\LARGE \bf
Terrain parameter estimation from proprioceptive sensing of the suspension dynamics in offroad vehicles 
}

\author{Jake Buzhardt \and Phanindra Tallapragada
\thanks{This work was supported by the Automotive Research Center (ARC), a US Army Center of Excellence for modeling and simulation of ground vehicles, under Cooperative Agreement W56HZV-19-2-0001 with the US Army DEVCOM Ground Vehicle Systems Center (GVSC).}
\thanks{
DISTRIBUTION A.  Approved for public release; distribution unlimited.
OPSEC \#: 5855 
}
\thanks{Jake Buzhardt and Phanindra Tallapragada are with the Department of Mechanical Engineering,
	Clemson	University, Clemson, SC, 29631, USA.
	\texttt{\small \{jbuzhar@g.clemson.edu, ptallap@clemson.edu\}}.
	}
}

\begin{document}

\maketitle
\thispagestyle{empty}
\pagestyle{empty}

\begin{abstract}
Offroad vehicle movement has to contend with uneven and uncertain terrain which present challenges to path planning and motion control for both manned and unmanned ground vehicles.  Knowledge of terrain properties can allow a vehicle  to  adapt  its  control  and  motion  planning  algorithms.  Terrain properties, however, can change on time scales of days or even hours, necessitating their online estimation. The kinematics and, in particular the oscillations experienced by an offroad vehicle carry a signature of the terrain properties. These terrain properties can thus be estimated from proprioceptive sensing of the vehicle dynamics with an appropriate model and estimation algorithm. In this paper, we show that knowledge of the vertical dynamics of a vehicle due to its suspension can enable faster and more accurate estimation of terrain parameters. The paper considers a five degree of freedom model that combines the well known half-car and bicycle models.
We show through simulation that the sinkage exponent, a parameter that can significantly influence the wheel forces from the terrain and thus greatly impact the vehicle trajectory, can be estimated from measurements of the vehicle's linear acceleration and rotational velocity, which can be readily obtained from an onboard IMU . 
We show that modelling the vertical vehicle dynamics can lead to significant improvement in both the estimation of terrain parameters and the prediction of the vehicle trajectory.  

\end{abstract}


\section{Introduction}
In many military applications, unmanned ground vehicles (UGVs) must navigate smoothly and efficiently over uneven, deformable terrain.  In such environments, unknown terrain properties greatly impact the vehicle's ability to track desired velocities or reference paths \cite{taheri2015technical,DALLAS202011,Dallas21_TerrainAdaptive}. Thus, it is necessary to consider the uncertainties in terrain properties in the planning and formulation of a control sequence to accomplish a desired task. While some nominal knowledge of terrain properties in an area can be known, terrain properties can change dramatically in a few days or even hours due to rain or snow and so the parameters of the terrain-vehicle interaction have to be estimated online.

In this paper, we show that by considering the vertical dynamics of the vehicle which have been neglected in previous works, we can gain a more complete understanding of the vehicle’s motion and interaction with the terrain. We extend previous efforts at online estimation of terrain parameters such as in \cite{DALLAS202011, antonov_vd_2011,TsiotrasUKF_ACC17}, by investigating the effects of varying terrain elevation, oscillations of the vehicle chassis, and forces within the vehicle suspension.  To this end, we develop a model that combines the classic half-car suspension model and the dynamic bicycle model.

To model the interaction of the tire with the soil, we use a Bekker-based wheel-soil interaction, or terramechanics, model as proposed in \cite{WongReece67,DALLAS202011,Ishigami_JFR07}. 
This model considers deformation of the terrrain surface and approximates the stress distributions in the wheel-soil contact region to predict the normal force and resistive forces on the wheel, which can be coupled into the vehicle dynamics model.
 
We show that small differences in the terrain parameters lead to significant differences in these forces, which in turn affect the trajectories of the vehicle states, especially the oscillations of the suspension and the vehicle body.
This contrast due to the terrain parameters can be exploited in order to estimate the parameters of the nonlinear terrain interaction model using an unscented Kalman filter (UKF).  

Previous works have also considered estimating parameters of the vehicle or terrain interaction model, but usually consider simpler models of the terrain or the vehicle. In \cite{DALLAS202011}, it was shown that parameters of the Bekker model can be estimated using a UKF, but the only vehicle model considered there was a dynamic bicycle with vertical dynamics neglected and the terrain was considered level.  In \cite{TsiotrasUKF_ACC17}, a UKF was also used to estimate vehicle and terrain parameters and vertical vehicle dynamics were considered, but the terrain was assumed to be level and non-deformable, with the terrain interaction being modelled by Pacejka's `magic formula' with unknown parameters. In \cite{PARK200441}, it was shown that variations in the terrain, specifically in the parameters of the Bekker model, can lead to significant variations in the vertical vehicle dynamics as the vehicle traverses an uneven terrain.  Here, we seek to exploit these differences to gain a better understanding of the vehicle-terrain interaction by modeling these vertical dynamics and estimating the unknown terrain parameters using a UKF. 

The results presented here have promising potential applications, extensions, and opportunities for future research.  
The ability
estimate unknown  model parameters could allow for the implementation of more sophisticated control strategies for agile yet safe maneuvers by UGVs.  
One interesting example of these are active suspensions, where the stiffness and damping properties of the vehicle suspension can be varied based on the estimated terrain parameters to better ensure safety and allow for more efficient and agile maneuvering. 

\section{Modelling} \label{sec:modelling}
The vehicle model considered here is formulated in two parts: a bicycle model to account for the steering and longitudinal dynamics and a half-car model to account for the vertical vehicle dynamics, including pitching and vertical oscillations of the the chassis.  We consider both a coupled version of these two models as well as a simplified model which neglects the vertical dynamics. In the coupled version of the model, the coupling is induced through the forces produced through wheel-terrain interactions, which are dependent on the normal reaction, which varies dynamically if the vertical vehicle dynamics are not neglected.  In the remainder of this section, these vehicle models as well as the terramechanics model are given, and a comparison of the models is presented.
\subsection{Vehicle Model}
  
\subsubsection{Bicycle Model}
\begin{figure}[t]
    \centering
    \scalebox{0.8}{
    \input{tikz_bicycle}
    }
    \caption{Bicycle model for longitudinal and steering dynamics }
    \label{fig:bicycle}
\end{figure}
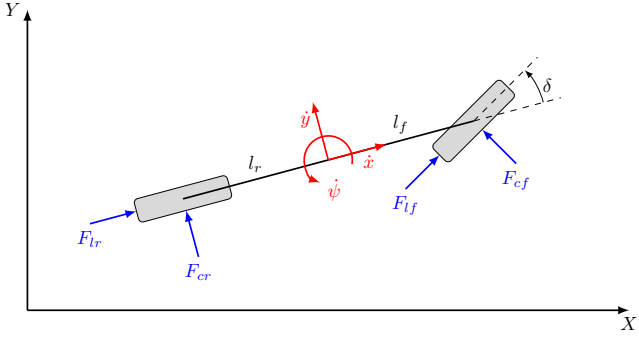

The steering and forward dynamics of the vehicle are modelled using a `bicycle model' \cite{rajamani,borelli_predictiveactivesteering07}, shown in Fig. \ref{fig:bicycle}, which considers longitudinal, lateral, and yaw degrees of freedom.  

The equations of motion are written by summing forces and moments in a body-fixed frame of reference, where $\dot{x}$ and $\dot{y}$ are the longitudinal and lateral velocities of the vehicle's center of mass in the body-fixed frame, and $\dot{\psi}$ is the yaw rate, as shown in Fig. \ref{fig:bicycle}:
\begin{align}
    m\ddot{x} & = m(\dot{y}\dot{\psi}\cos\theta + \dot{z}\dot{\theta}) + F_{xf} + F_{xr} + F_u\label{eq:bicycle1} \\
    & =m(\dot{y}\dot{\psi}\cos\theta + \dot{z}\dot{\theta}) + F_{lf}\cos\delta - F_{cf}\sin\delta + F_{lr} + F_u\nonumber\\
     m\ddot{y} & = m(\dot{z}\dot{\psi}\sin\theta-\dot{x}\dot{\psi}\cos\theta) + F_{yf} + F_{yr} \label{eq:bicycle2} \\
    & =m(\dot{z}\dot{\psi}\sin\theta - \dot{x}\dot{\psi}\cos\theta)  + F_{lf}\sin\delta + F_{cf}\cos\delta + F_{cr}\nonumber\\
     I_{z} \ddot{\psi} &= F_{yf}l_f - F_{yr}l_r  \label{eq:bicycle3} \\
    & = (F_{lf}\sin\delta + F_{cf}\cos\delta)l_f - F_{cr}l_r \nonumber 
\end{align}
These equations are also dependent upon the steering angle $\delta$, which is assumed to be commanded instantaneously; the vehicle mass $m$ and yaw moment of inertia $I_z$; the distances from the wheels to the center of gravity $l_f$ and $l_r$;  and the wheel forces $F_{lf}$, $F_{cf}$, $F_{lr}$, $F_{cr}$, and $F_u$.  Following \cite{borelli_predictiveactivesteering07}, here the subscripts $(\cdot)_f$ and $(\cdot)_r$ indicate the front and rear wheel, respectively, while the subscripts $(\cdot)_l$ and $(\cdot)_c$ represent longitudinal and cornering (lateral) directions relative to the wheel (e.g. $F_{lf}$ is the longitudinal force at the front wheel).  These tire forces are given by the Bekker deformable terrain model, explained in Section \ref{sec:terramechanics}.  $F_u$ is an additional forcing applied for actuation of the vehicle.  

Equations (\ref{eq:bicycle1})-(\ref{eq:bicycle3}) can be integrated to obtain the velocity of the center of mass, which it is convenient to represent in terms of the global reference frame, denoted by $X$-$Y$.  
\begin{align}
    \dot{X} &= \dot{x}\cos\psi - \dot{y}\sin\psi\\
    \dot{Y} &= \dot{x}\sin{\psi} + \dot{y}\cos\psi
\end{align}
In simulating this bicycle model throughout the rest of this paper, the state vector is taken to be $\begin{bmatrix} X,Y,\psi,\dot{x}, \dot{y}, \dot{\psi} \end{bmatrix}^{\intercal}$.

\subsubsection{Half Car Model}
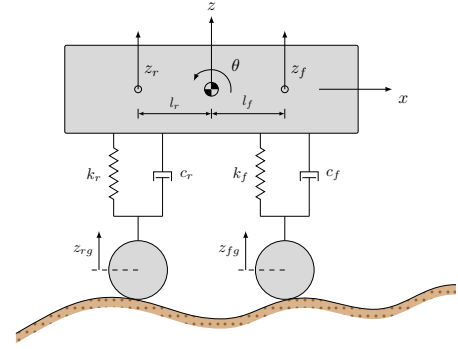
\begin{figure}[t]
    \centering
    \scalebox{0.65}{
    \input{tikz_halfcar}
    }
    \caption{Schematic of half-car model for vertical dynamics}
    \label{fig:half-car}
\end{figure}

The vertical and pitching dynamics of the vehicle are modelled using a `half-car' model \cite{rajamani}, shown in Fig. \ref{fig:half-car}, which considers the stiffness and damping of the front and rear suspensions of a vehicle traversing over uneven terrain. 

The equations of motion for the half-car are derived by summing forces and moments in the vertical frame, where the coordinates considered are the pitch angle $\theta$ and the vertical displacement $z$ of the center of mass of the vehicle, measured from static equilibrium. 
\begin{align}
     m\ddot{z} ~=~ &-k_r(z_r - z_{rg}) - c_r(\dot{z}_r - \dot{z}_{rg}) \label{eq:HalfCar1} \\ 
     & \qquad - k_f(z_f - z_{fg}) - c_f(\dot{z}_f - \dot{z}_{fg}) \nonumber
    \\
    I_y\ddot{\theta} ~=~ &\big(k_r(z_r - z_{rg})l_r + c_r(\dot{z}_r - \dot{z}_{rg})l_r \label{eq:HalfCar2} \\
    & \qquad - k_f(z_f - z_{fg})l_f -c_f(\dot{z}_f - \dot{z}_{fg})l_f\big) \cos\theta\nonumber
\end{align}
The vehicle parameters introduced here are the stiffnesses $k$ and damping constants $c$ at the front and rear suspensions and the pitching moment of inertia $I_y$.  The intermediate measurements $z_f$ and $z_r$ are the vertical translations at the front and rear axles, which are given by 
\begin{align*}
    z_f = z + l_f\sin\theta \qquad \text{and} \qquad 
    z_r = z-l_r\sin\theta
\end{align*}
The variables $z_{fg}$ and $z_{rg}$ represent the vertical displacements of the ground from static equilibrium at the front and rear wheels. These are found from the terrain elevation profile and the sinkage of the wheel into the deformable terrain as 
\(
z_g = H(X,Y) - h_f.
\)
The terrain elevation profile $H(X,Y)$ is taken to be a smooth, continuous, known function of the global coordinates,  and the computation procedure for the sinkage of the wheel, $h_f$ is discussed in Section \ref{sec:terramechanics}.

In order to compute the forces from the Bekker terrain interaction model, the normal reaction at each of the wheels must also be known.  Assuming that each wheel remains in contact with the ground ($\ddot{z}_g = \ddot{H}$), the dynamic normal reaction is calculated as 
\begin{equation}
    N = \frac{1}{2}mg - k(z - z_g) - c(\dot{z} - \dot{z}_g) + m_w\ddot{H}
\end{equation}
where the front and rear wheel subscripts are included as appropriate, and $m_w$ is the mass of the wheel.

\subsection{Terrain Interaction} \label{sec:terramechanics}
\begin{figure}[t]
    \centering
    \scalebox{0.8}{
    \input{tikz_terrain}  
    }
    \caption{Schematic for model of wheel-terrain interaction}
    \label{fig:terramechanics}
\end{figure}
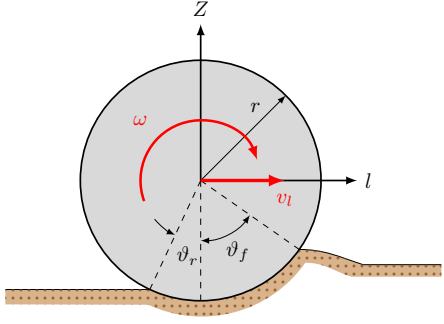
In order to compute the forces $F_{l}$ and $F_{c}$ at the wheel resulting from the terrain interactions, a Bekker-based terramechanics model is implemented.  With such, the forces are computed by integrating the normal and shear stress distributions over the contact region between the wheel and terrain.  
The normal and shear stress distributions $\sigma$, $\tau_x$, and $\tau_y$ are taken to be functions of the sinkage $h$, which, in turn, is assumed to be a function of the contact angle $\vartheta$. These are given by 
\begin{align}
    \sigma(\vartheta) &= \left(\frac{k_c}{b} + k_{\phi} \right)  h(\vartheta)^n \label{eq:sigma_stress}\\ 
    \tau_x(\vartheta) &= 
    \left(c + \sigma(\vartheta)\tan\phi \right)
    \left(1-e^{-j_x/k_x} \right)\\
    \tau_y(\vartheta) &= 
    \left(c + \sigma(\vartheta)\tan\phi \right)
    \left(1-e^{-j_y/k_y} \right)
\end{align}
where $k_c$, $k_{\phi}$, and $n$ are terrain parameters, $b$ is the effective width of the wheel, and the sinkage $h$ is given as follows.
\begin{equation*}
    h(\vartheta) = 
    \begin{cases}
    r\left( \cos\vartheta - \cos\vartheta_f\right) & \vartheta_m\leq\vartheta\leq\vartheta_f\\
    r\left( \cos\vartheta_e - \cos\vartheta_f\right) \qquad \qquad& \vartheta_r\leq\vartheta\leq\vartheta_m\\
    \end{cases}
\end{equation*}
Here $\vartheta_f$ and $\vartheta_r$ are the front and rear contact angles as depicted in Fig. \ref{fig:terramechanics}, and $\vartheta_m$ is the angle of maximum normal stress.  These computed as  
\begin{align}
\vartheta_f &= \cos^{-1}\left(1 - \frac{h_f}{r}\right)\\[1ex]
    \vartheta_m &= (a_0 + a_1s)\,\vartheta_f\\[1ex]
    \vartheta_r &= (b_0 + b_1s)\,\vartheta_f\\[1ex]
    \vartheta_e & = \vartheta_f - \left(\frac{\vartheta - \vartheta_r}{\vartheta_m - \vartheta_r}\right)(\vartheta_f - \vartheta_m)
\end{align}
where $a_0$, $a_1$, $b_0$, $b_1$ are soil-dependent parameters. 
For the shear stresses $\tau_x(\vartheta)$ and $\tau_y(\vartheta)$, the shear displacements $j_x(\vartheta)$ and $j_y(\vartheta)$ are also needed. These are given by 
\begin{align}
    j_x(\vartheta) &= 
    r\big[(\vartheta_f - \vartheta) - (1-s)(\sin\vartheta_f - \sin\vartheta)\big]\\[1ex]
    j_y(\vartheta) &= 
    r(1-s)(\vartheta_f-\vartheta)\cdot\tan\beta
\end{align}
where $s = (r\omega - v_l)/r\omega$ is the slip ratio and $\beta = \tan^{-1}(v_c/v_l)$ is the side-slip angle, which can be represented in terms of the vehicle velocities $\dot{x}$, $\dot{y}$, $\dot{\psi}$ and steering angle $\delta$ as 
\[
\begin{split}
\tan\beta_f &= 
\frac{(\dot{y}+l_f\dot{\psi})\cos\delta - \dot{x}\sin\delta}{(\dot{y}+l_f\dot{\psi})\sin\delta + \dot{x}\cos\delta}
\\[1ex]
\tan\beta_r &=
\frac{(\dot{y}-l_r\dot{\psi})}{\dot{x}}
\end{split}
\]
for the front and rear wheels respectively.  Since wheel rotation $\omega$ is not tracked as a state in our simulations, the slip ratio $s$ is assumed to maintain a constant value.

\begin{figure*}[h]
    \centering
    \begin{subfigure}[b]{0.36\linewidth}
        \centering
        \includegraphics[width = \linewidth]{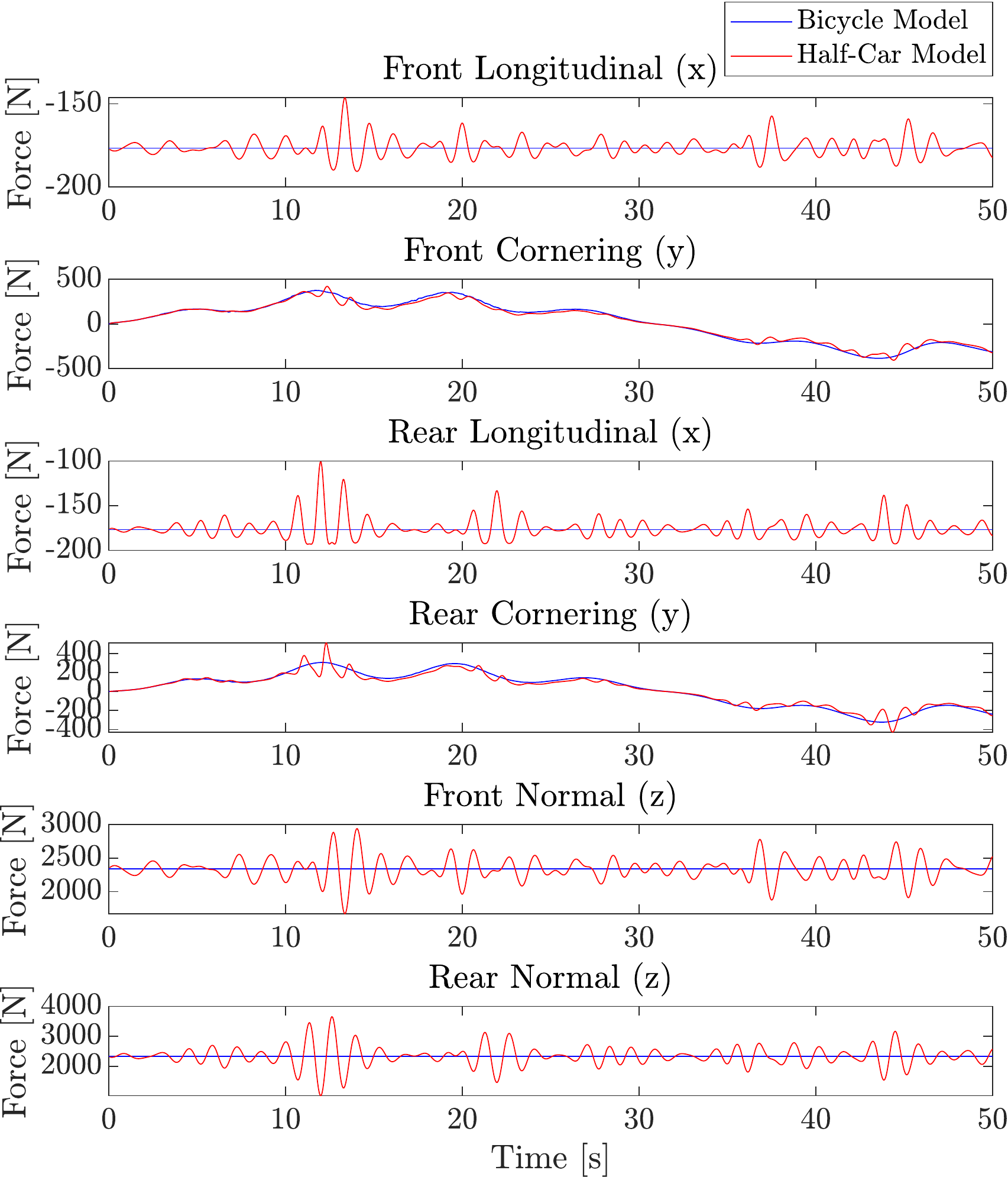}
        \caption{}\label{fig:force_comp_forces}
    \end{subfigure}
    \begin{subfigure}[b]{0.36\linewidth}
        \centering
        \includegraphics[width = \linewidth]{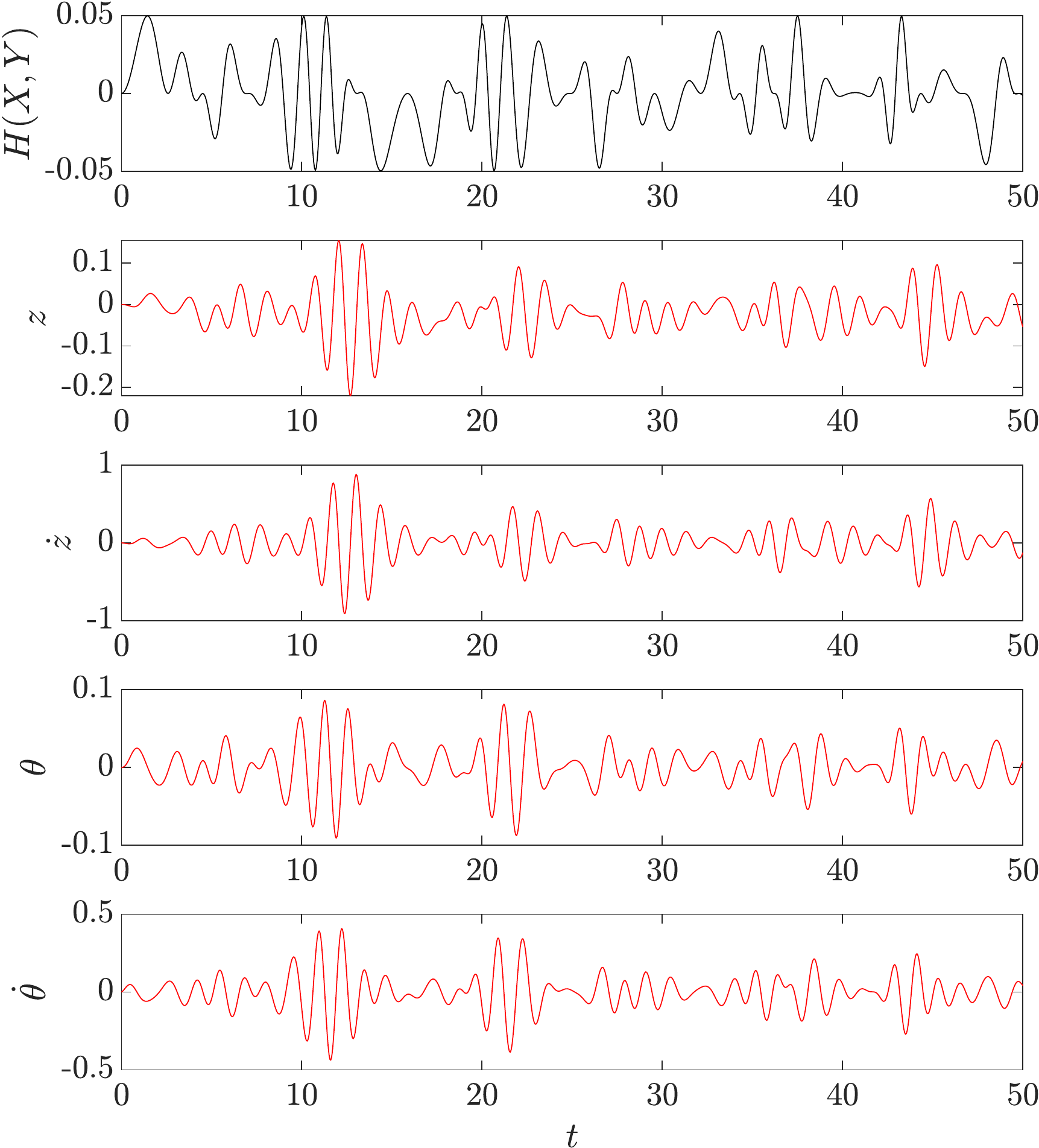}
        \caption{}\label{fig:force_comp_vert}
    \end{subfigure}
    \begin{subfigure}[b]{0.24\linewidth}
        \centering
        \includegraphics[width = \linewidth]{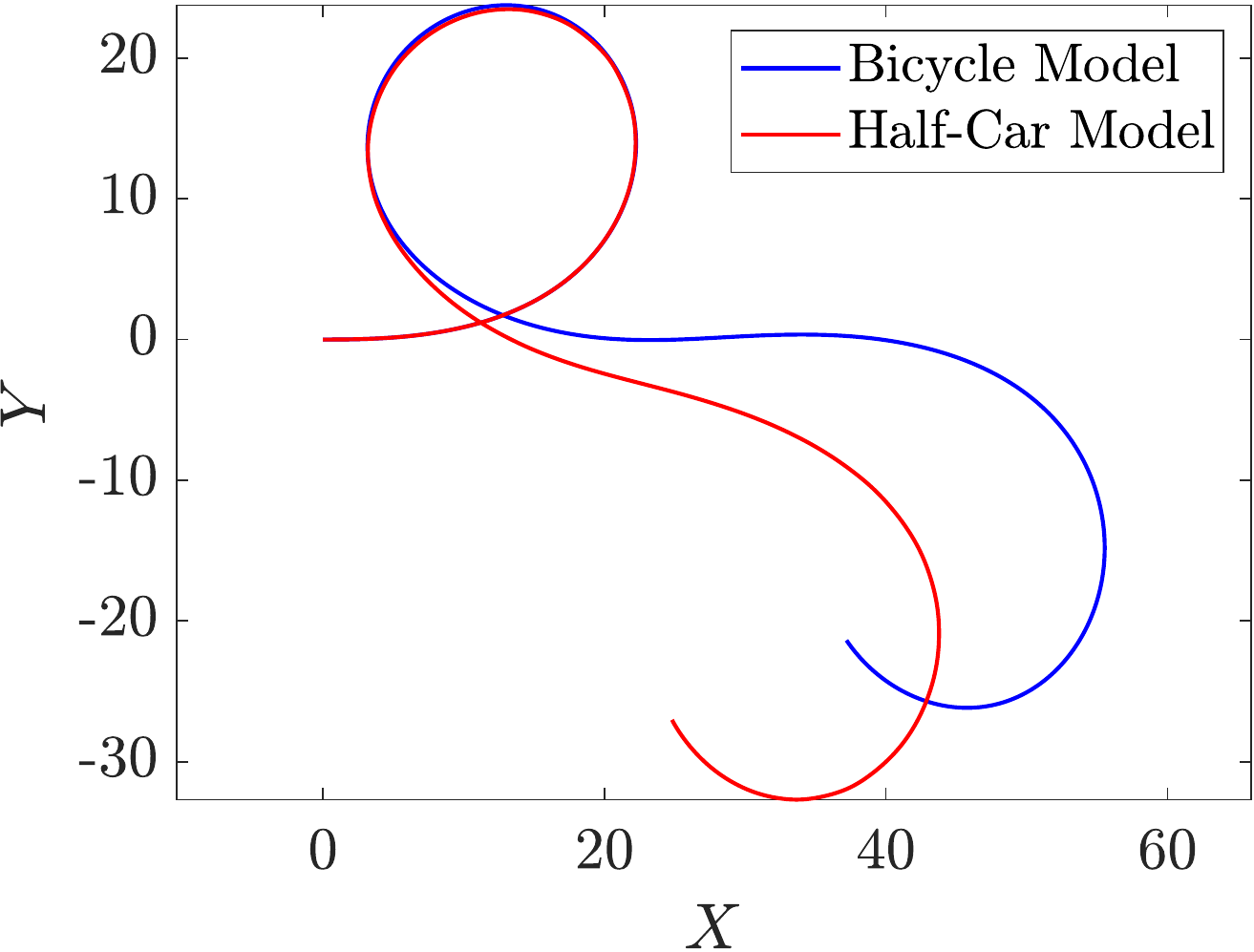}
        \caption{}\label{fig:force_comp_trajXY}
        \includegraphics[width = \linewidth]{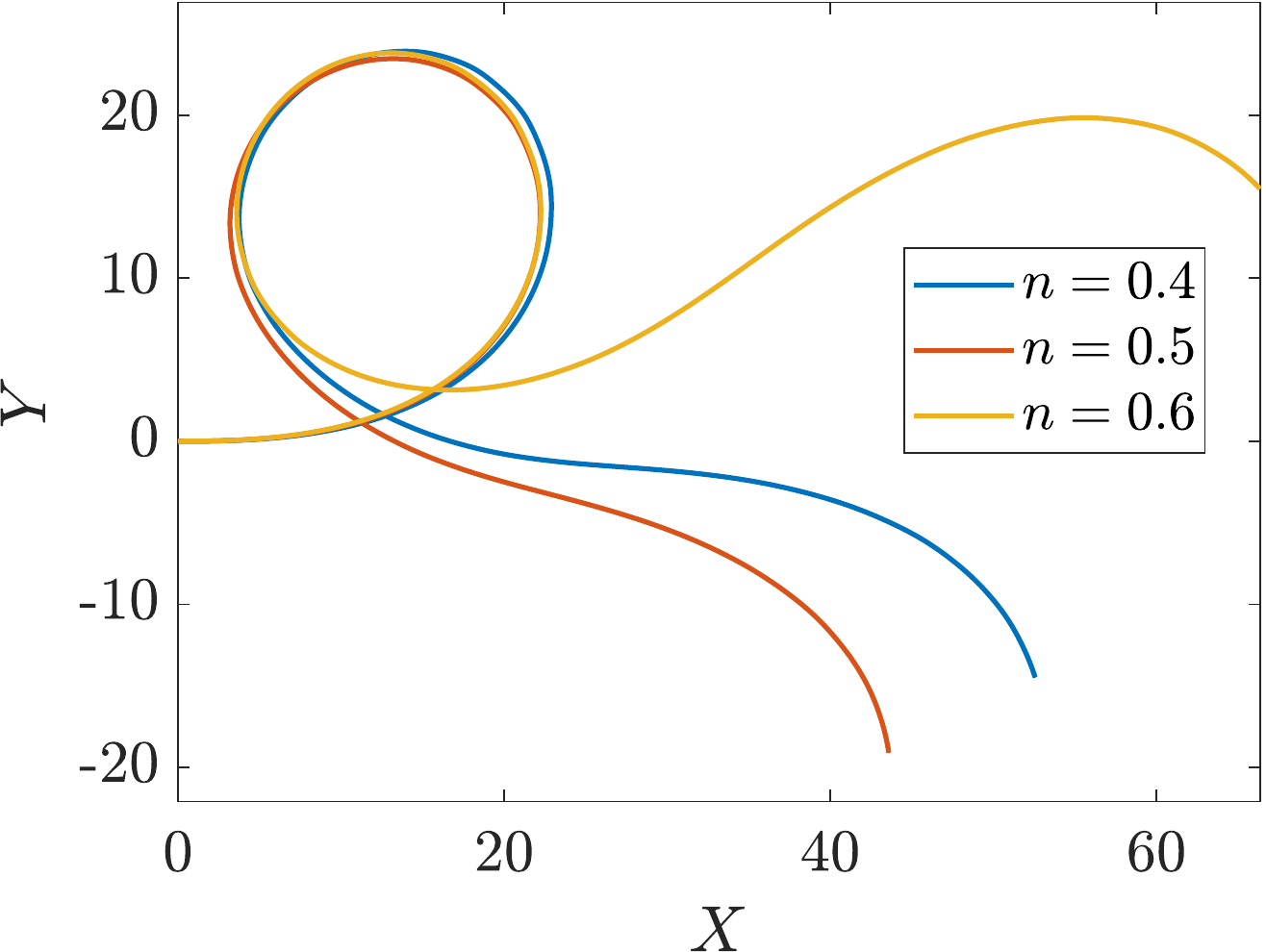}
        \caption{} \label{fig:traj_comp}
    \end{subfigure}
    \caption{Comparison of the (a) force outputs at the wheel from the Bekker model (b) terrain roughness $H(X,Y)$, suspension and pitch kinematics and (c) the corresponding spatial trajectories of the half-car model and bicycle model.  
    Subfigure (d) gives a comparison of spatial trajectories with varying sinkage exponent, $n$.
    }
    \vspace{-1em}
    \label{fig:force_comp}
\end{figure*}

With this, $\sigma(\vartheta)$, $\tau_x(\vartheta)$, and $\tau_y(\vartheta)$ are fully defined with the exception of the maximum sinkage $h_f$.  These distributions can then be integrated to yield the forces on the wheel as 
\begin{align}
    F_x &= \int_{\vartheta_r}^{\vartheta_f}rb\left(\tau_x(\vartheta)\cos\vartheta - \sigma(\vartheta)\sin\vartheta \right)d\vartheta \label{eq:Fx}\\[1ex]
    F_y &= -\int_{\vartheta_r}^{\vartheta_f} rb\tau_y(\vartheta) d\vartheta\label{eq:Fy}\\[1ex]
    F_z &= \int_{\vartheta_r}^{\vartheta_f}rb\left(\tau_x(\vartheta)\sin\vartheta + \sigma(\vartheta)\cos\vartheta \right)d\vartheta\label{eq:Fz}
\end{align}
Finally, the maximum sinkage $h_f$ can be found by applying the boundary condition that the resultant force in the vertical direction is equal to the normal reaction.  In this work, $h_f$ is found using Eq. \ref{eq:Fz} as the zero of the expression $F_z - N = 0$ through a Newton-Raphson iterative procedure. Once $h_f$ is found, the stress distributions are fully defined and Eqs. \ref{eq:Fx}-\ref{eq:Fy} can be used to compute the tractive and lateral forces on the wheel.

\subsection{Vehicle Model Comparison} \label{sec:modelcomp}
In order to compare the bicycle model (Eqs.(\ref{eq:bicycle1}-\ref{eq:bicycle3})) to the half-car model (Eqs. (\ref{eq:bicycle1}-\ref{eq:bicycle3},\ref{eq:HalfCar1}-\ref{eq:HalfCar2})), we examine the force outputs from the Bekker-model for each model and the corresponding effect on the vehicle trajectory. 
As mentioned in Section \ref{sec:modelling}, the coupling between the vertical dynamics and dynamics of the bicycle model in the half-car model is through the dependence of the tractive and cornering forces on the normal reaction, which dynamically varies due to the displacement of the sprung mass of the half-car. In the bicycle model alone, these vertical dynamics are neglected and the normal force used to compute the wheel forces is simply the static normal reaction. If the vehicle is traversing level ground, these two models yield the same result, as the sprung mass will not be displaced.  But in many applications, it is necessary for a vehicle to traverse an uneven terrain, in which case the normal reaction will vary, which can lead to significantly different trajectories.  To see this, we consider a terrain elevation profile specified by the following function in terms of the global position coordinates
\begin{equation}
H(X,Y) = H_0\sin^2(0.5\,X)\cos(1.5\,Y) \label{eq:H_profile}
\end{equation}
with $H_0 = 0.05$ and where the output of this function is the terrain elevation in meters. 
For the Bekker model, we use the terrain parameters corresponding to clay \cite{WongReece67}.  For the purpose of this simulation, we apply a sinusoidal longitudinal forcing $ F_u(t) = m(0.8 + 0.5\sin0.8t)$ and steering angle $ \delta(t) = 0.3\sin(0.1 t)$.

This simulation demonstrates the effect of the vertical vehicle dynamics on the forces at the wheel due to the terrain interaction and, in turn, on the vehicle trajectory.
Fig. \ref{fig:force_comp_forces} shows normal forces 
along with the resulting tractive and cornering forces at each wheel for a 50s simulation.  As expected, the normal reaction is held constant throughout the simulation for the bicycle model, but for the half-car model, the normal reaction varies with displacements of the sprung mass due to changes in the terrain elevation.  This leads to smooth, slowly changing wheel forces at each wheel for the bicycle model, but more rapid oscillations in the tractive forces in the half-car model.  Fig. \ref{fig:force_comp_vert} shows the vertical dynamics from the simulation, which lead to the discrepancy, as they are neglected in the bicycle model.  The impact of this on the spatial trajectory of the vehicle in the $XY$ frame is shown in Fig. \ref{fig:force_comp_trajXY} . We see that the error due to the differences in the forcings accumulates over time, leading to differences of as much as 15m between the predicted trajectory from the bicycle model and from the half-car model. 

Furthermore, this simulation framework also allows us to consider the sensitivity of the vehicle's dynamics to the terrain parameters. It has been shown \cite{DALLAS202011}, that the sinkage exponent $n$ has the most significant impact of all of the terrain parameters on the output of the Bekker model. Using the same prescribed trajectories of steering and longitudinal velocities, and the terrain parameters corresponding to clay, as before, we run simulations of the vehicle motion over the uneven using the coupled half-car model with different sinkage exponents.  The trajectories shown in Fig. \ref{fig:traj_comp} correspond to sinkage exponents ranging from 0.4 to 0.6 (held constant during each individual simulation).  

These results motivate two lines of research. Fig. \ref{fig:force_comp} shows that it is important in at least some cases to consider the vertical dynamics, as neglecting them can lead to a miscalculation of the forces at the wheel, which can lead to accumulation of error in the predicted vehicle trajectory.  Fig. \ref{fig:traj_comp} shows that it is crucial to have good estimates of the deformable terrain parameters, especially the sinkage exponent, in order to perform common control tasks, such as tracking a desired trajectory.  These sorts of error build-ups also could become significant in predictive control frameworks, which rely on accurate predictions from the model in order to formulate optimal control strategies. 

\section{Parameter Estimation Formulation} \label{sec:UKF}
Motivated by the results of Section \ref{sec:modelcomp}, we now seek to implement a parameter estimation strategy, as it is clear that good estimates of the terrain parameters should enable better predictions of the vehicle motion on the deformable terrain. For this, we implement an UKF \cite{WanVanDerMerwe_ukf_2000,julier1997new,thrun2005probabilistic} in a formulation specifically tailored to parameter estimation, as shown in \cite{vanDerMerwe_thesis} and outlined briefly below. 

For the parameter estimation problem, we model the dynamics of the unknown parameters as a stationary process driven by Gaussian noise and define a new discrete-time state space representation 
\begin{align}
    w_{k+1} &= w_k + n_k\\
    d_k &= f(\xi_k, w_k) + e_k
\end{align}
where $w$ represents a vector of unknown parameters; $f(\cdot)$ is a nonlinear mapping of states, $\xi$ and parameters to an observation vector $d$; and $n_k$ and $e_k$ are Gaussian process and observation noises, respectively.

The filter is initialized with an initial estimate of the parameters $\hat{w}_0$ and the parameter covariance $P_{w_0}$.  At each timestep $k$, the a priori estimate of the parameter mean and covariance are updated as
\begin{align}
    \hat{w}_k^- &= \hat{w}_{k-1}\\
    P_{w_k}^- &= P_{w_{k-1}} + R_{n}.
\end{align}
With this, a set of $2L+1$ \emph{sigma points} are distributed about the current parameter estimate as 
\begin{equation}
    \mathcal{W}_{k|k-1} = 
    \left[ 
    \hat{w}_k^- ~,~ 
    \hat{w}_k^- \pm \sqrt{(L+\lambda)P_{w_k}^-} 
    \right]
\end{equation}
where $L$ is the length of the parameter vector and 
\(
\lambda = \alpha^2(L+\kappa) - L
\)
where $\alpha$ and $\kappa$ are scaling parameters to adjust the distribution of sigma points \cite{vanDerMerwe_thesis}.  Each of the sigma points are then used to propagate the observation model forward 
\begin{equation}
    \mathcal{D}_{k|k-1} = f(\xi_k,\mathcal{W}_{k|k-1})
\end{equation}
and the results are used to compute an a priori observation as a weighted sum of the predicted observations from the sigma points.
\begin{equation}
 \hat{d}_k^- = \sum_{i=0}^{2L}a_i^{(m)}\mathcal{D}_{i,k|k-1}
 \end{equation}
The observation and observation-parameter covariances are then updated as 
\begin{align*}
    P_{d_k} &= 
    R_{e_k} + 
    \sum_{i=0}^{2L} a_i^{(c)} \,
    (\mathcal{D}_{i,k|k-1} - \hat{d}_k^-)(\mathcal{D}_{i,k|k-1} - \hat{d}_k^-)^{\intercal}
    \\
    P_{w_kd_k} &=
    \sum_{i=0}^{2L}a_i^{(c)}\,
     (\mathcal{W}_{i,k|k-1} - \hat{w}_k^-)(\mathcal{D}_{i,k|k-1} - \hat{d}_k^-)^{\intercal}
\end{align*}
where the weights $a^{(m)}$ and $a^{(c)}$ are 
\[
\begin{split}
    &a_0^{(m)} = \frac{\lambda}{L+\lambda}
    \qquad ,\qquad a_0^{(c)} = \frac{\lambda}{L+\lambda} - \alpha^2 + 3 \,,\\[1ex]
    &a_i^{(m)} = a_i^{(c)} = \frac{1}{2(L+\lambda)} \qquad \text{for } i=1, \dots, 2L
\end{split}
\]
and with this, the Kalman gain is 
\begin{equation}
    K_k = P_{w_k}P_{d_k}^{-1}.
\end{equation}

Once an observation $d_k$ is received, the a posteriori mean and covariance of the parameters are updated as follows.
\begin{align}
    \hat{w}_k &= \hat{w}_k^- + K_k(d_k - \hat{d}_k^-)\\[1ex]
    P_{w_k} &= P_{w_k}^- - K_kP_{d_k}K_k^\intercal
\end{align}

This formulation of the UKF is useful, as it separates the state estimation problem from the parameter estimation problem.  The tunable parameters of the filter include the sigma point scaling and weighting parameters $\alpha$ and $\kappa$, as well as the covariance of the artificial process noise, $R_n$.  The covariance $R_n$ determines the allowed magnitude of changes in the parameter values at each time step.  So, choosing too large of a value can lead to instability, while choosing too small of a value can lead to stagnation or slow convergence. For a full derivation and discussion of the filter, the reader is referred to Ref. \cite{vanDerMerwe_thesis}.  For the results presented herein, $\kappa$ is set to zero and the parameters $\alpha$ and $R_n$ are chosen via a parameter sweep, with parameters selected to minimize the mean square error in the estimate.  The same set of filter parameters are used in all of the results shown for each respective model.

\section{Results}
The unscented Kalman filter for parameter estimation shown in Section \ref{sec:UKF} is applied for the purpose of estimating terrain parameters.  It has been shown that, of all the terrain parameters in the Bekker terrain interaction model, the force outputs are most sensitive to the sinkage exponent, $n$, introduced in Eq. \ref{eq:sigma_stress} \cite{DALLAS202011}. Therefore, we assume that all other parameters are known or can be easily measured, and focus our efforts on estimating the sinkage exponent alone.  For a ground truth simulation, we use the half-car model described by Eqs. (\ref{eq:bicycle1}-\ref{eq:bicycle3},\ref{eq:HalfCar1}-\ref{eq:HalfCar2}) with a fixed set of vehicle and terrain parameters. The vehicle parameters are chosen to be representative of a Polaris MRZR D2 \cite{polaris_mrzr} and the terrain parameters for the Bekker model are used for clay and sand from Ref. \cite{WongReece67}.  We compare the estimation capability of the bicycle model (Eqs. (\ref{eq:bicycle1}-\ref{eq:bicycle3})), which neglects the vertical vehicle dynamics, to that of the half-car model (Eqs. (\ref{eq:bicycle1}-\ref{eq:bicycle3},\ref{eq:HalfCar1}-\ref{eq:HalfCar2})). We consider observations of the linear accelerations in the $x$, $y$, and $z$ directions measured in a body-fixed reference frame and angular velocities about the $y$ and $z$ axes of the vehicle body.  For the bicycle model alone, the vertical acceleration and pitch velocity are unmodeled and therefore unobserved in the filter. The observations from the ground truth simulation are corrupted with Gaussian noise with standard deviations of $0.2$ m/s$^2$ in the accelerations and $0.0175$ rad/s$^2$ in the angular velocities before being transmitted to the filter.
The longitudinal forcing is prescribed as  $F_u = m(0.8+0.5\sin(0.8\,t))$ for the simulations on clay and $F_u = m(1.8+0.6\sin(0.8\,t))$ for the simulations on sand.  The terrain elevation profile used for the estimation is the same as in Eq. \ref{eq:H_profile}
and is assumed to be fully known in the model.  
\begin{figure}[t]
\centering
 \begin{subfigure}[t]{0.49\linewidth}
        \centering
        \includegraphics[width = \linewidth]{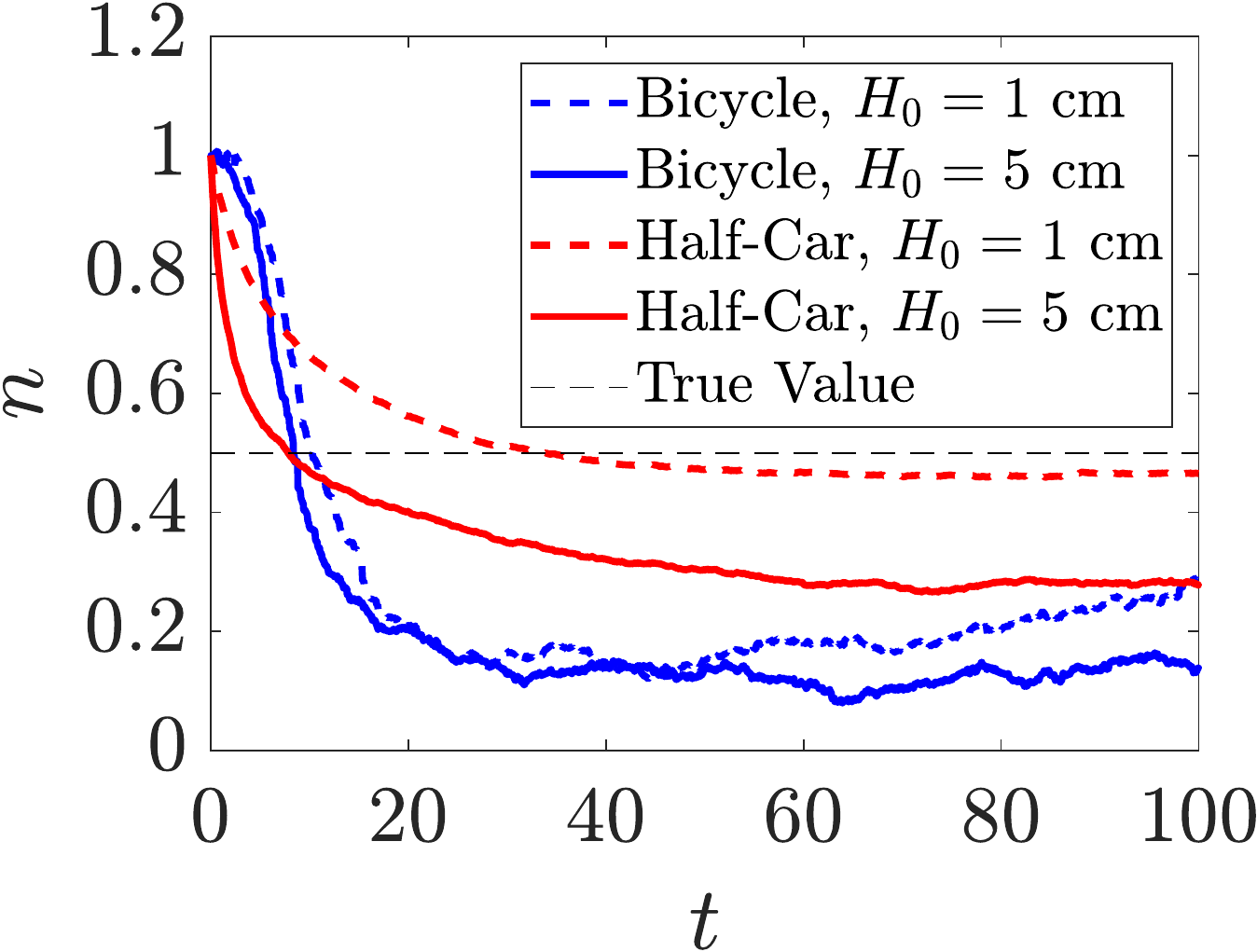}
        \caption{}\label{fig:ukf_clay1}
    \end{subfigure}
    \hfill
    \begin{subfigure}[t]{0.49\linewidth}
        \centering
        \includegraphics[width = \linewidth]{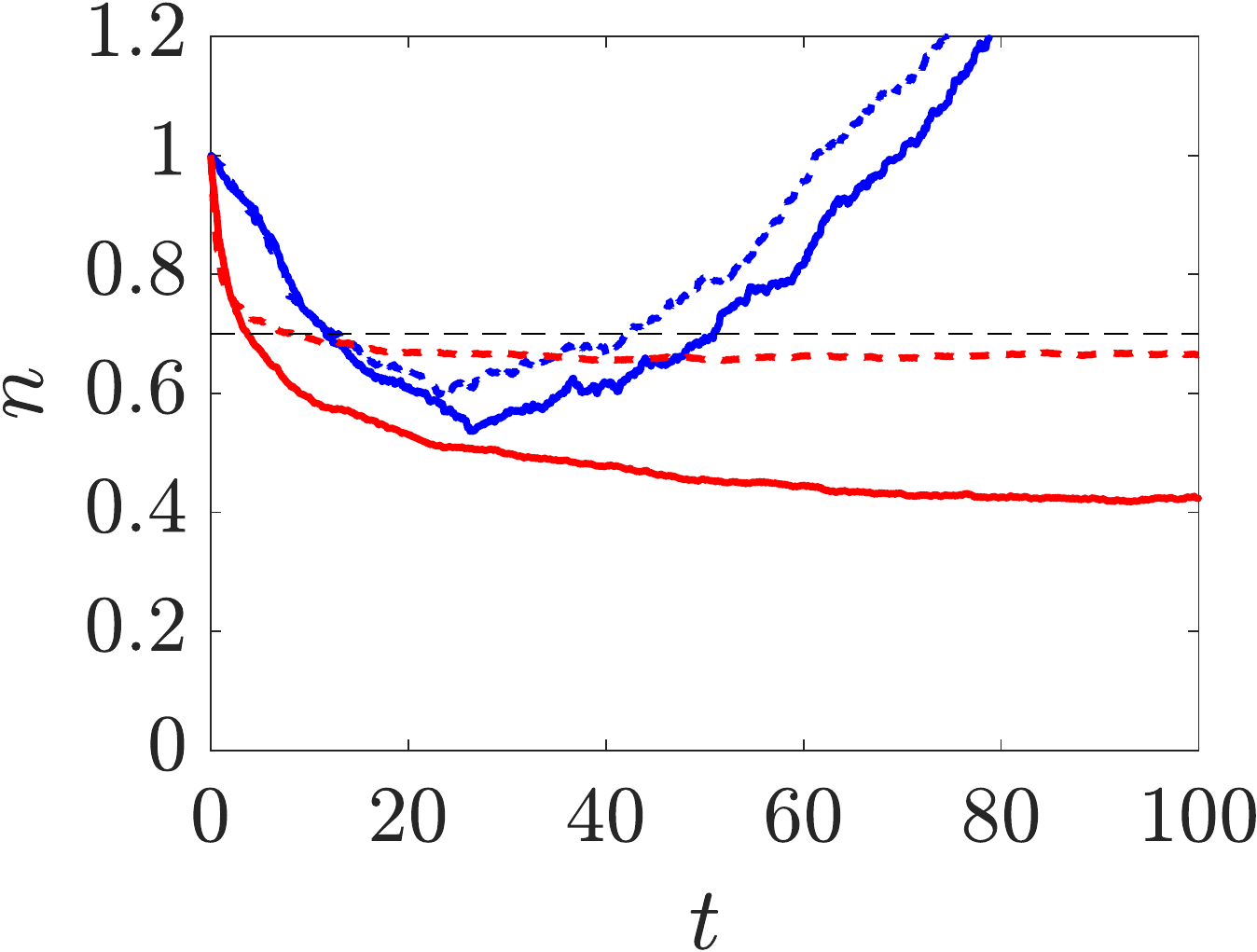}
        \caption{}\label{fig:ukf_sand1}
    \end{subfigure}
    \\
     \begin{subfigure}[t]{0.49\linewidth}
        \centering
        \includegraphics[width = \linewidth]{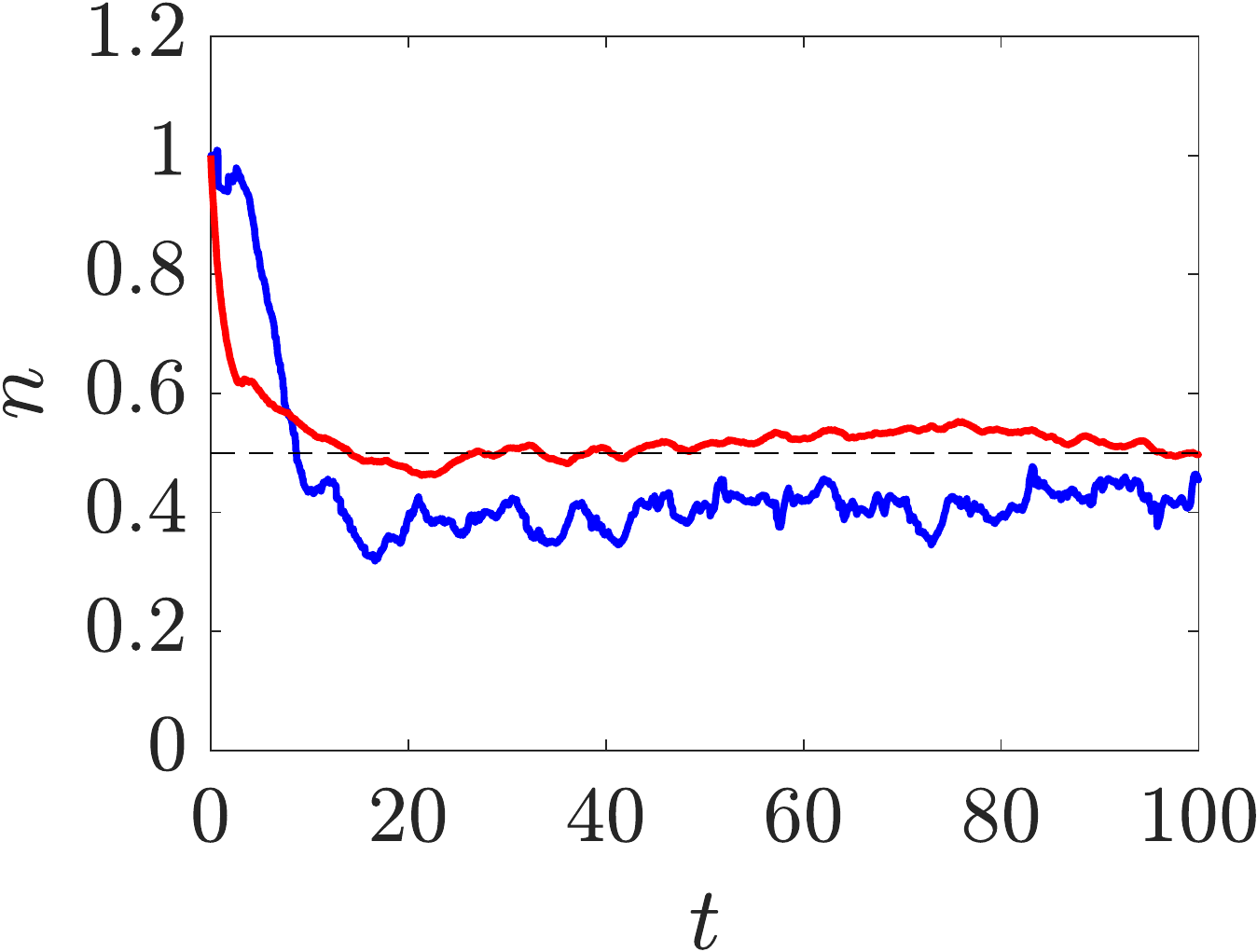}
        \caption{}\label{fig:ukf_clay2}
    \end{subfigure}
    \hfill
    \begin{subfigure}[t]{0.49\linewidth}
        \centering
        \includegraphics[width = \linewidth]{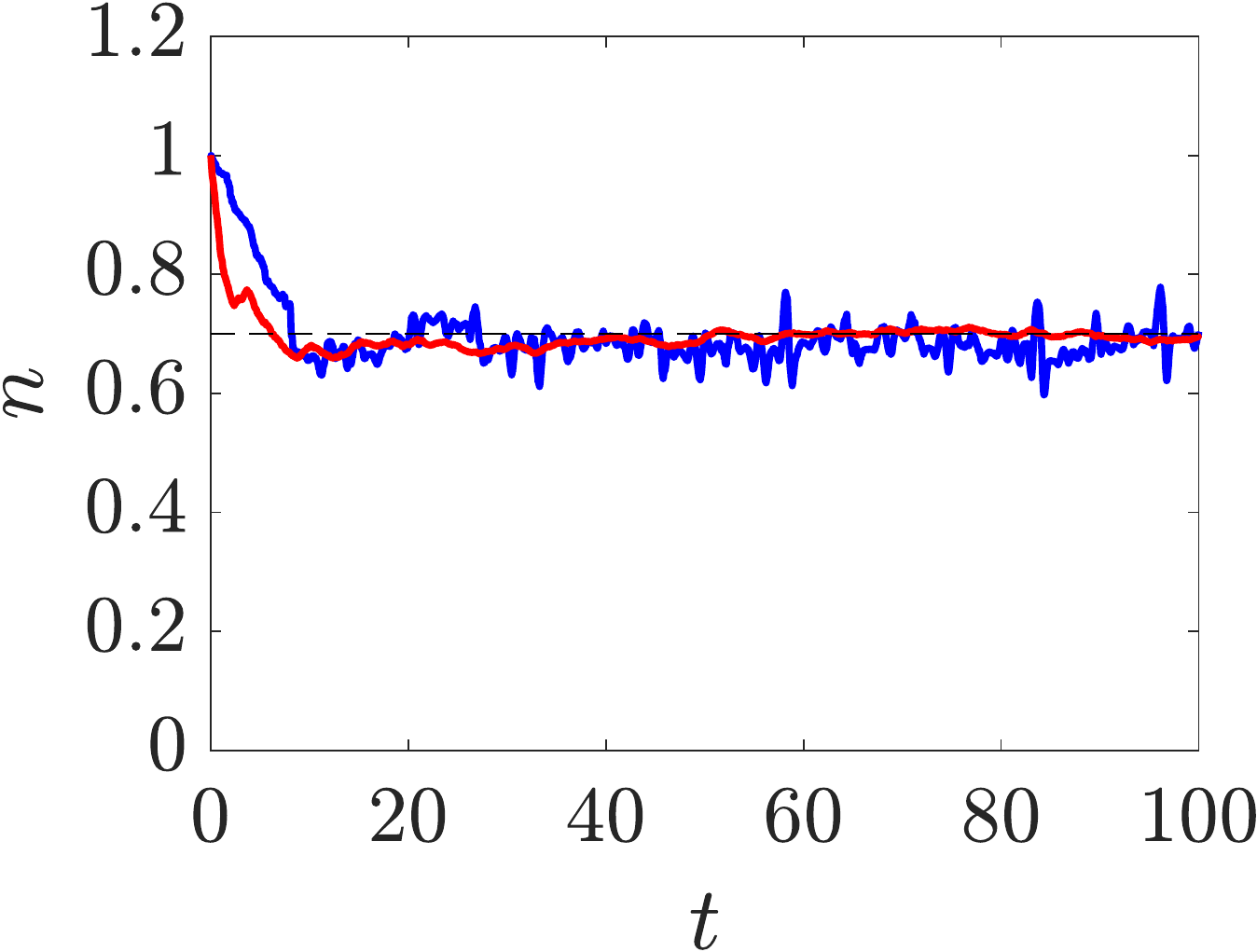}
        \caption{}\label{fig:ukf_sand2}
    \end{subfigure}
    \\
     \begin{subfigure}[t]{0.49\linewidth}
        \centering
        \includegraphics[width = \linewidth]{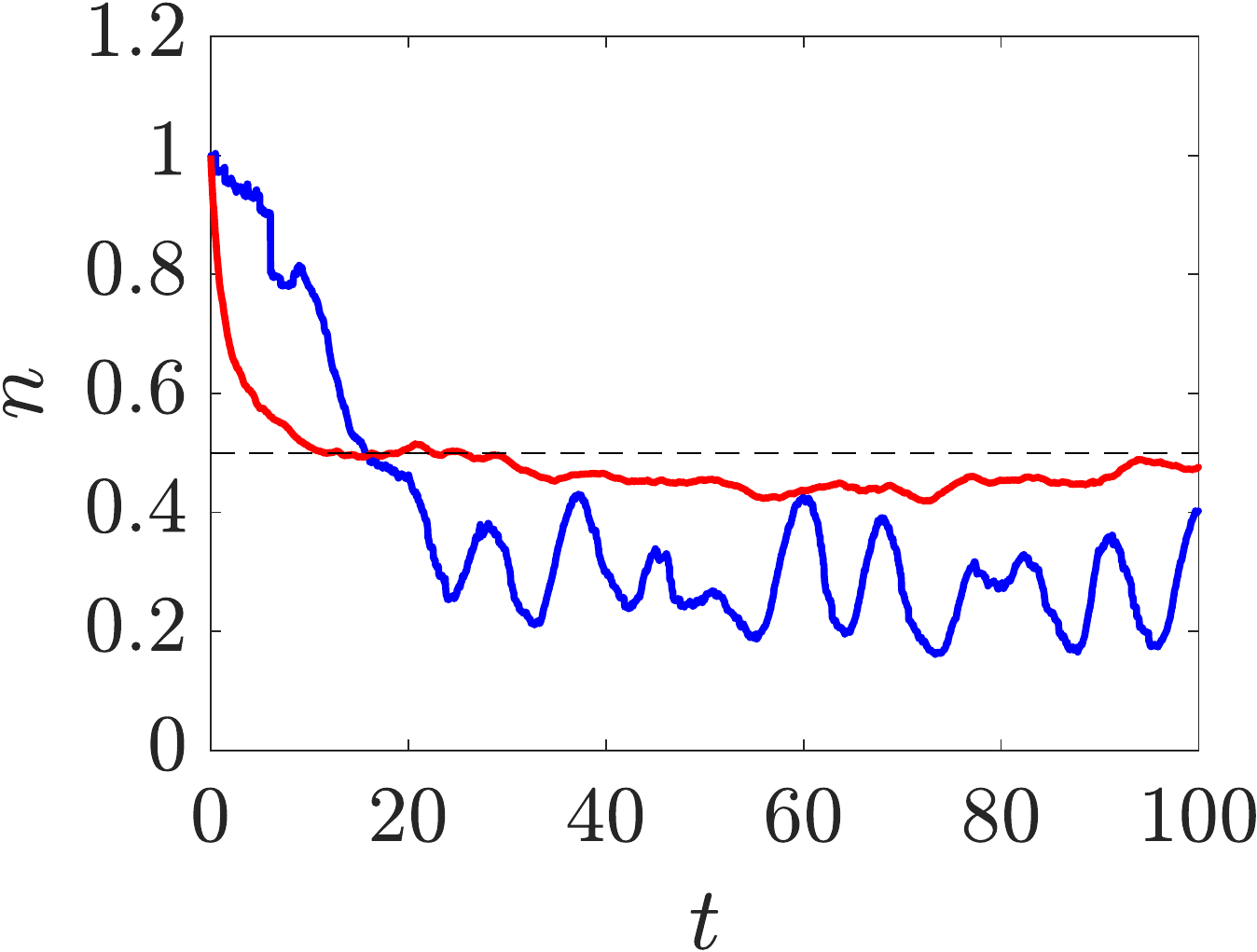}
        \caption{}\label{fig:ukf_clay3}
    \end{subfigure}
    \hfill
    \begin{subfigure}[t]{0.49\linewidth}
        \centering
        \includegraphics[width = \linewidth]{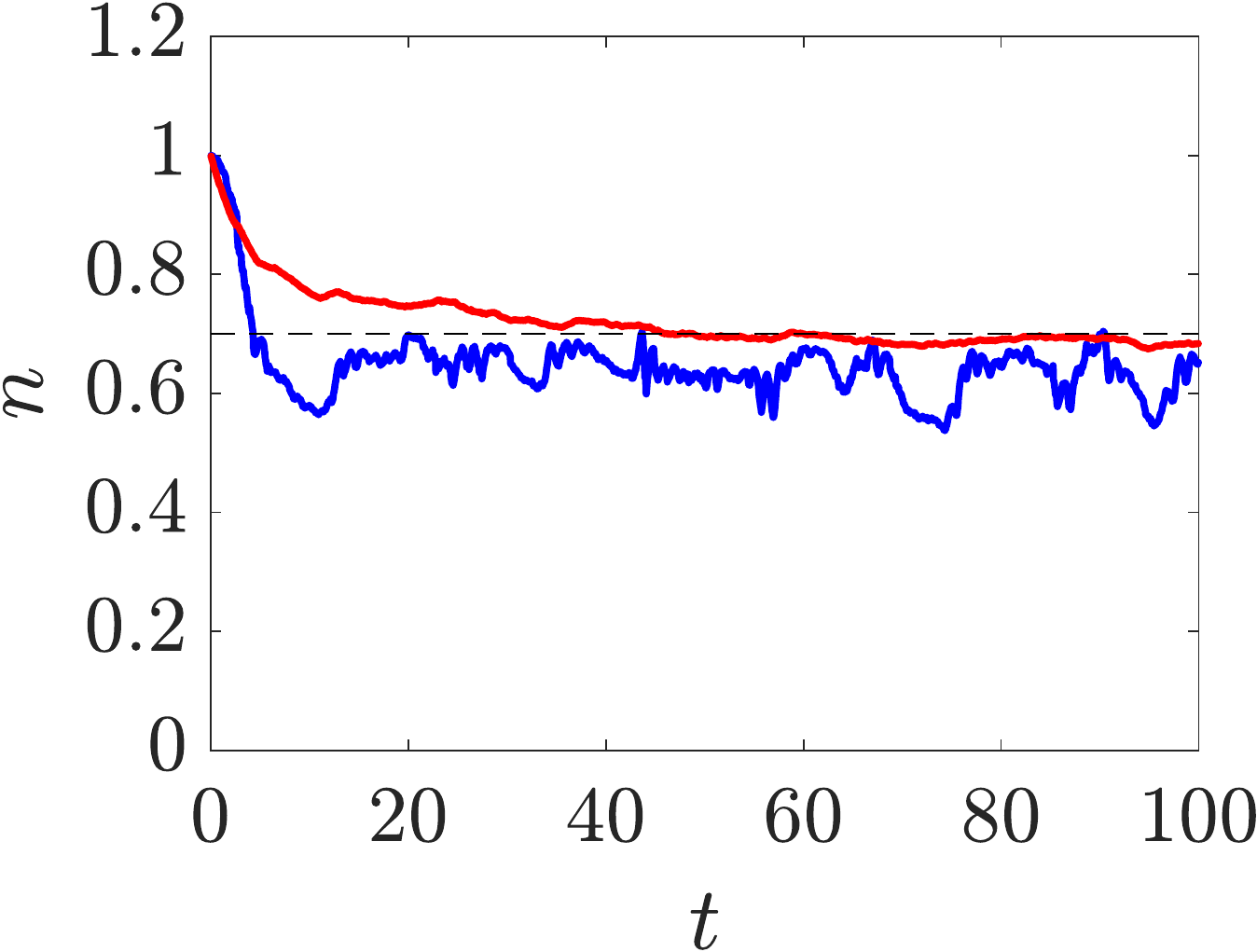}
        \caption{}\label{fig:ukf_sand3}
    \end{subfigure}
    \caption{Estimation results for the sinkage exponent, $n$ using the Bicycle model and Half-car model.  The initial guess for each case was chosen as $n=1.0$.  The steering, soil, and error properties associated with these simulations is summarized in Table \ref{tab:results}. 
    } 
    \label{fig:results}
    \vspace{-1.5em}
\end{figure}

\begin{table}[!b]
        \vspace{-1.5em}
\caption{Terrain, steering, forcing, and estimation error corresponding to the plots in Fig. \ref{fig:results}.}
    \resizebox{\columnwidth}{!}{%
    \begin{tabular}{c  c  c  c  c  c}
    \hline \hline
    \multicolumn{4}{c}{} & \multicolumn{2}{c}{Mean Square Error}
    \\
    Fig. &  Soil & $H_0$ [m] & Steering, $\delta(t)$ [rad] &  Half-Car & Bicycle \\
    \hline  \hline
    \ref{fig:ukf_clay1} & Clay & 0.05 & $0$ &  3.47e-2 & 1.26e-1 \\
    \ref{fig:ukf_clay1} & Clay & 0.01 & $0$ &  1.08e-2 & 9.81e-2 \\
    \ref{fig:ukf_clay2} & Clay & 0.05 & $0.2\sin(t)$&  3.04e-3 & 2.02e-2 \\    \ref{fig:ukf_clay3} & Clay & 0.05 & $0.5\sin(0.3\,t)$&  4.26e-3 & 5.78e-2\\[0.5ex]
    \ref{fig:ukf_sand1} & Sand & 0.05 & $0$ &  5.26e-2 & 1.55e-1\\
    \ref{fig:ukf_sand1} & Sand & 0.01 & $0$ &  1.66e-3 & 1.97e-1\\
    \ref{fig:ukf_sand2} & Sand & 0.05 & $0.2\sin(t)$ &  9.13e-4 & 3.53e-3\\
    \ref{fig:ukf_sand3} & Sand & 0.05 & $0.5\sin(0.3\,t)$ &  3.16e-3 & 7.11e-3\\
    \hline \hline
    \end{tabular}
    }
    \label{tab:results}
\end{table}

We consider three different steering inputs:  zero steering (Figs. \ref{fig:ukf_clay1}, \ref{fig:ukf_sand1}), a small but rapidly oscillating steering angle (Figs. \ref{fig:ukf_clay2}, \ref{fig:ukf_sand2}), and a wide, but slowly oscillating steering angle (Figs. \ref{fig:ukf_clay3}, \ref{fig:ukf_sand3}).  These inputs are summarized in Table \ref{tab:results} along with the mean square error of the sinkage exponent from its true value over a 100 second simulation.  Estimation results are also depicted in Fig. \ref{fig:results}.  We see that in each of the cases considered, the estimation results using the half-car model, both in terms of the mean-square error, and the speed of convergence. 
For the case of zero steering shown in Figs. \ref{fig:ukf_clay1} and \ref{fig:ukf_sand1}, neither estimator converges to the true value of $n$ on the terrain with elevation $H_0=0.05$ m. This is likely due to the fact that no lateral forcings are induced, and thus the parameters must be estimated from the longitudinal and vertical dynamics alone. However, we see that on terrain with elevation $H_0 = 0.01$ m, the half-car estimator converges quite closely to the true value, while there is significant error in the bicycle model's estimate.  This indicates that in some cases, the vertical and longitudinal dynamics are sufficient to estimate the terrain parameters with the half-car model, while the bicycle model requires some lateral dynamics for good parameter estimation.

\section{Conclusion}
The online estimation of terrain parameters is necessary for agile and safe driving of offroad UGVs or even manned vehicles. This paper presents a 5-degree of freedom vehicle model that uses proprioceptive sensing measurements and a UKF to obtain faster and more accurate estimates of the sinkage exponent on an uneven terrain.  The possibility of fast online estimation of terrain parameters that using better vehicle dynamic models opens up the possibility for using active suspensions and other controllers for vibration isolation for onboard cameras, online path replanning and correction and safe and agile operation of vehicles in uncertain terrains.

\addtolength{\textheight}{-10cm}   




\bibliographystyle{IEEEtran}
\bibliography{gvsc}

\end{document}

%% file: tikz_bicycle.tex
\begin{tikzpicture}

\coordinate (O) at (0,0);
\coordinate (F) at (15:2.5);
\coordinate (R) at (195:2.5);
\coordinate (OO) at (-5,-2.5);
\coordinate (OX) at (5,-2.5);
\coordinate (OY) at (-5,2.5);

\filldraw[fill = gray!30,shift={(15:2.5)}, rotate=45,scale = 0.2,rounded corners = 2pt] (-4,-1) rectangle (4,1);

\filldraw[fill = gray!30,shift={(195:2.5)},rotate=15,scale = 0.2,rounded corners = 2pt] (-4,-1) rectangle (4,1);
\draw[thick] (O)--(F) node [midway,anchor=south,scale = 0.9] {$l_f$};
\draw[thick] (O)--(R) node [midway,anchor=south,scale = 0.9] {$l_r$};
\draw[dashed] (F)--(15:4.0);
\draw[dashed,rotate around={30:(F)}] (F)--(15:4.0);
\draw[-latex] (F)+(15:1.2) arc (15:45:1.2) node[midway,right,scale=0.9] {$\delta$};
\draw[thick,red, -latex] (O) -- ++(15:1);
\draw[thick,red, -latex] (O) -- ++(105:1);
\node[scale = 0.9,red,anchor=north] at (15:0.7) {$\dot{x}$};
\node[scale = 0.9,red,anchor=east] at (105:0.7) {$\dot{y}$};
\draw[thick,red, -latex] (O)+(-10:0.4) arc (-10:250:0.4);
\node[scale = 0.9,red] at (0.1,-0.5) {$\dot{\psi}$};
\draw[thick,blue,latex-] (R)+(-75:0.2) -- ++(-75:1) node[scale=0.9,anchor=north] {$F_{cr}$};
\draw[thick,blue,latex-] (R)+(-165:0.8) -- ++(-165:1.6) node[scale=0.9,anchor=north] {$F_{lr}$};
\draw[thick,blue,latex-] (F)+(-45:0.2) -- ++(-45:1) node[scale=0.9,anchor=north] {$F_{cf}$};
\draw[thick,blue,latex-] (F)+(-135:0.8) -- ++(-135:1.6) node[scale=0.9,anchor=north] {$F_{lf}$};

\draw[thick,-latex] (OO)--(OX) node[scale = 0.9,anchor=north] {$X$};
\draw[thick,-latex] (OO)--(OY) node[scale = 0.9,anchor=east] {$Y$};

\end{tikzpicture}

%% file: tikz_halfcar.tex
\begin{tikzpicture}
\filldraw[fill = gray!30, rounded corners = 1pt] (-3,-0.9) rectangle (3,0.9);
\draw(1.5,0) circle [radius = 2pt];
\draw(-1.5,0) circle [radius = 2pt];
\draw[spring](-2.0,-0.9)-- ++(0,-1.7)node[midway,xshift=-0.4cm,scale=0.9] {$k_r$};
\draw[damper] (-1.0,-0.9)-- ++(0,-1.7) node[midway,xshift=0.5cm,scale=0.9] {$c_r$} ;
\draw(-2.0,-2.6) -- (-1.0,-2.6);
\draw(-1.5,-2.6) -- +(0,-0.5);
\filldraw[fill = gray!30] (-1.5,-3.7) circle [radius=0.6];

\draw[spring](1.0,-0.9)-- ++(0,-1.7) node[midway,xshift=-0.4cm,scale=0.9] {$k_f$};
\draw[damper] (2.0,-0.9)-- ++(0,-1.7) node[midway,xshift=0.5cm,scale=0.9] {$c_f$}; ;
\draw(1.0,-2.6) -- (2.0,-2.6);
\draw(1.5,-2.6) -- +(0,-0.5);
\filldraw[fill = gray!30] (1.5,-3.7) circle [radius=0.6];

\draw[dashed] (1.5,-3.7) -- ++(-1,0);
\draw[-latex] (0.7,-3.7) -- ++(0,0.8) node[midway, anchor=east,scale = 0.9] {$z_{fg}$};

\draw[dashed] (-1.5,-3.7) -- ++(-1,0);
\draw[-latex] (-2.3,-3.7) -- ++(0,0.8) node[midway, anchor=east,scale = 0.9] {$z_{rg}$};

\def\mypath {(-4,-5) to [out=30,in=170]  
      (-1.6,-4.3) to [out=-10,in=180] 
      (0.2,-4.7) to[out=0,in = 190] 
      (1.6,-4.3) to [out = 10,in = 175] 
      (3.2,-4.5) to [out = -5,in = 180] 
      (5.0,-4.0) -- 
      (5.0,-4.2) to [out = -180,in = -5]
      (3.2,-4.7) to [in = 10,out = 175]
      (1.6,-4.5) to[in=0,out = 190]
      (0.2,-4.9) to [in=-10,out=180]
      (-1.6,-4.5) to [in=30,out=170]
      (-4,-5.2) --cycle}

\fill[brown!60] \mypath;
\pattern[pattern = dots, pattern color = brown!80!black] \mypath;

\draw (-4,-5) to [out=30,in=170]  
      (-1.6,-4.3) to [out=-10,in=180] 
      (0.2,-4.7) to[out=0,in = 190] 
      (1.6,-4.3) to [out = 10,in = 175] 
      (3.2,-4.5) to [out = -5,in = 180] 
      (5.0,-4.0) 
      ;

\draw[-latex] (0,0) -- (0,1.5) node[anchor = south] {$z$};
\draw[-latex] (1.5,0) -- (1.5,1.2) node[midway, anchor = north west] {$z_f$}; 
\draw[-latex] (-1.5,0) -- (-1.5,1.2) node[midway, anchor = north west] {$z_r$}; 

\draw[-latex] (2.2,0) -- ++(1.5,0) node[anchor = north west] {$x$}; 

\draw[ultra thin,latex-latex] (-1.5,-0.5) -- (0,-0.5) node[midway, anchor = south,scale = 0.8] {$l_r$};
\draw[ultra thin,latex-latex] (1.5,-0.5) -- (0,-0.5) node[midway, anchor = south,scale = 0.8] {$l_f$};
\draw[ultra thin] (0,-0.4) -- (0,-0.6);
\draw[ultra thin] (-1.5,-0.4) -- (-1.5,-0.6);
\draw[ultra thin] (1.5,-0.4) -- (1.5,-0.6);

\fill[radius = 0.4em] (0,0) -- ++(0.4em,0) arc [start angle=0,end angle=90] -- ++(0,-0.8em) arc [start angle=270, end angle=180];
\draw (0,0) circle [radius = 0.4em];

\draw[-latex] (0,0)+(-10:0.4) arc (-10:155:0.4);
\node at (45:0.7) {$\theta$};
\end{tikzpicture}

%% file: tikz_terrain.tex
\begin{tikzpicture}[scale = 2]
    \coordinate (O) at (0,0);
    \coordinate (f) at (-35:1);
    \coordinate (r) at (-115:1);
    \coordinate (ur) at (0.7071,0.7071);
    \coordinate (B) at (0,-1);
    \coordinate (eh) at (1.35,-0.7);

    \def\mypath {($(r) + (-1.2,0)$) -- (r) arc (-115:-35:1) to [out = 0, in = 162] (eh) -- (2,-0.7) --  (2,-0.8) -- ($(eh) + (-0.1,-0.1)$) to [out = 162, in = 0]($(f)+ (0.05,-0.11)$)  arc (-35:-115:1.05) -- (-1.6226,-1.03) -- cycle};
    \fill[brown!60] \mypath;
    \fill[gray!30] (O) circle (1);
    \draw[dashed] (O)--(f);
    \draw[dashed] (O)--(r);
    \draw[-latex] (O)--(ur);
    \node[xshift = -0.5cm, yshift = -0.2cm] at (ur)  {$r$};

    \draw (f) to [out = 0, in = 162] (eh);
    \draw (eh) -- (2,-0.7);
    \draw (r) -- ($(r) + (-1.2,-0)$);
    \draw[dashed] (O) -- (B);
    \draw[latex-latex] (0,-0.5) arc (-90:-35:0.5);
    \node at (-62.5:0.5) [below,xshift = 1ex] {$\vartheta_f$};
    \node at (-102.5:0.5) [below] {$\vartheta_r$};
    \draw[-latex] (-140:0.5) arc (-140:-115:0.5);
    
    \draw[-latex,thick] (O) -- (1.3,0) node[right] (x) {$l$};
    \draw[-latex,thick] (O) -- (0,1.3) node[above] (z) {$Z$};
    \draw[-latex, ultra thick, red] (0,0) -- (0.7,0) node[below, yshift = -0.5ex] {$v_l$};
    \draw[-latex, very thick, red] (-160:0.5) arc (-160:-340:0.5);
    \node at (-0.5,0.5) {$\color{red}\omega$};

    \pattern[pattern = dots, pattern color = brown!80!black]          \mypath;
    \draw[thick] (O) circle (1);
    

\end{tikzpicture}

%% file: bt_estimation_acc2021.bbl
\begin{thebibliography}{10}
\providecommand{\url}[1]{#1}
\csname url@samestyle\endcsname
\providecommand{\newblock}{\relax}
\providecommand{\bibinfo}[2]{#2}
\providecommand{\BIBentrySTDinterwordspacing}{\spaceskip=0pt\relax}
\providecommand{\BIBentryALTinterwordstretchfactor}{4}
\providecommand{\BIBentryALTinterwordspacing}{\spaceskip=\fontdimen2\font plus
\BIBentryALTinterwordstretchfactor\fontdimen3\font minus
  \fontdimen4\font\relax}
\providecommand{\BIBforeignlanguage}[2]{{%
\expandafter\ifx\csname l@#1\endcsname\relax
\typeout{** WARNING: IEEEtran.bst: No hyphenation pattern has been}%
\typeout{** loaded for the language `#1'. Using the pattern for}%
\typeout{** the default language instead.}%
\else
\language=\csname l@#1\endcsname
\fi
#2}}
\providecommand{\BIBdecl}{\relax}
\BIBdecl

\bibitem{taheri2015technical}
S.~Taheri, C.~Sandu, S.~Taheri, E.~Pinto, and D.~Gorsich, ``A technical survey
  on terramechanics models for tire--terrain interaction used in modeling and
  simulation of wheeled vehicles,'' \emph{Journal of Terramechanics}, vol.~57,
  pp. 1--22, 2015.

\bibitem{DALLAS202011}
\BIBentryALTinterwordspacing
J.~Dallas, K.~Jain, Z.~Dong, L.~Sapronov, M.~P. Cole, P.~Jayakumar, and
  T.~Ersal, ``Online terrain estimation for autonomous vehicles on deformable
  terrains,'' \emph{Journal of Terramechanics}, vol.~91, pp. 11--22, 2020.
  [Online]. Available: \url{https://doi.org/10.1016/j.jterra.2020.03.001}
\BIBentrySTDinterwordspacing

\bibitem{Dallas21_TerrainAdaptive}
J.~Dallas, M.~P. Cole, P.~Jayakumar, and T.~Ersal, ``Terrain adaptive
  trajectory planning and tracking on deformable terrains,'' \emph{IEEE
  Transactions on Vehicular Technology}, pp. 1--1, 2021.

\bibitem{antonov_vd_2011}
S.~Antonov, A.~Fehn, and A.~Kugi, ``Unscented kalman filter for vehicle state
  estimation,'' \emph{Vehicle System Dynamics}, vol.~49, no.~9, p. 1497–1520,
  2011.

\bibitem{TsiotrasUKF_ACC17}
C.~You and P.~Tsiotras, ``Vehicle modeling and parameter estimation using
  adaptive limited memory joint-state ukf,'' in \emph{2017 American Control
  Conference (ACC)}, 2017, pp. 322--327.

\bibitem{WongReece67}
\BIBentryALTinterwordspacing
J.-Y. Wong and A.~Reece, ``Prediction of rigid wheel performance based on the
  analysis of soil-wheel stresses part i. performance of driven rigid wheels,''
  \emph{Journal of Terramechanics}, vol.~4, no.~1, pp. 81--98, 1967. [Online].
  Available: \url{https://doi.org/10.1016/0022-4898(67)90105-X}
\BIBentrySTDinterwordspacing

\bibitem{Ishigami_JFR07}
\BIBentryALTinterwordspacing
G.~Ishigami, A.~Miwa, K.~Nagatani, and K.~Yoshida, ``Terramechanics-based model
  for steering maneuver of planetary exploration rovers on loose soil,''
  \emph{Journal of Field Robotics}, vol.~24, no.~3, pp. 233--250, 2007.
  [Online]. Available: \url{https://doi.org/10.1002/rob.20187}
\BIBentrySTDinterwordspacing

\bibitem{PARK200441}
\BIBentryALTinterwordspacing
S.~Park, A.~Popov, and D.~Cole, ``Influence of soil deformation on off-road
  heavy vehicle suspension vibration,'' \emph{Journal of Terramechanics},
  vol.~41, no.~1, pp. 41--68, 2004. [Online]. Available:
  \url{https://doi.org/10.1016/j.jterra.2004.02.010}
\BIBentrySTDinterwordspacing

\bibitem{rajamani}
R.~Rajamani, \emph{Vehicle dynamics and control}.\hskip 1em plus 0.5em minus
  0.4em\relax Springer Science \& Business Media, 2011.

\bibitem{borelli_predictiveactivesteering07}
P.~Falcone, F.~Borrelli, J.~Asgari, H.~E. Tseng, and D.~Hrovat, ``Predictive
  active steering control for autonomous vehicle systems,'' \emph{IEEE
  Transactions on control systems technology}, vol.~15, no.~3, pp. 566--580,
  2007.

\bibitem{WanVanDerMerwe_ukf_2000}
E.~Wan and R.~Van Der~Merwe, ``The unscented kalman filter for nonlinear
  estimation,'' in \emph{Proceedings of the IEEE 2000 Adaptive Systems for
  Signal Processing, Communications, and Control Symposium (Cat. No.00EX373)},
  2000, pp. 153--158.

\bibitem{julier1997new}
S.~J. Julier and J.~K. Uhlmann, ``New extension of the kalman filter to
  nonlinear systems,'' in \emph{Signal processing, sensor fusion, and target
  recognition VI}, vol. 3068.\hskip 1em plus 0.5em minus 0.4em\relax
  International Society for Optics and Photonics, 1997, pp. 182--193.

\bibitem{thrun2005probabilistic}
W.~B. Sebastian~Thrun and D.~Fox, \emph{Probabilistic Robotics}.\hskip 1em plus
  0.5em minus 0.4em\relax The MIT Press, 2005.

\bibitem{vanDerMerwe_thesis}
R.~van~der Merwe, ``Sigma-point kalman filters for probabilistic inference in
  dynamic state-space models,'' Ph.D. dissertation, OGI School of Science \&
  Engineering, 2004.

\bibitem{polaris_mrzr}
\BIBentryALTinterwordspacing
\emph{Specs: Polaris MRZR D2 - Military Tan}.\hskip 1em plus 0.5em minus
  0.4em\relax Polaris Government \&; Defense. [Online]. Available:
  \url{https://military.polaris.com/en-us/mrzr-d2-military-tan/specs/}
\BIBentrySTDinterwordspacing

\end{thebibliography}
